\documentclass[nohyperref]{article}

\usepackage{microtype}
\usepackage{graphicx}
\usepackage{subfig}
\usepackage{booktabs} 

\usepackage{hyperref}

\usepackage[accepted]{icml2023}

\usepackage{amsmath}
\usepackage{amssymb}
\usepackage{mathtools}
\usepackage{amsthm}
\usepackage[capitalize,noabbrev]{cleveref}

\theoremstyle{plain}
\newtheorem{theorem}{Theorem}[section]

\newtheorem{lemma}[theorem]{Lemma}

\theoremstyle{definition}

\theoremstyle{remark}

\usepackage[textsize=tiny]{todonotes}

\icmltitlerunning{Hybrid Quantum-Classical Hadamard Transform Layer}

\begin{document}

\twocolumn[
\icmltitle{A Hybrid Quantum-Classical Approach based on the Hadamard Transform for the Convolutional Layer }



\icmlsetsymbol{equal}{*}

\begin{icmlauthorlist}
\icmlauthor{Hongyi Pan}{yyy}
\icmlauthor{Xin Zhu}{yyy}
\icmlauthor{Salih Atici}{yyy}
\icmlauthor{Ahmet Enis Cetin}{yyy}
\end{icmlauthorlist}

\icmlaffiliation{yyy}{Department of Electrical and Computer Engineering, University of Illinois Chicago, Chicago, IL, USA}

\icmlcorrespondingauthor{Hongyi Pan}{hpan21@uic.edu}
\icmlcorrespondingauthor{Ahmet Enis Cetin}{aecyy@uic.edu}

\icmlkeywords{Machine Learning, ICML}

\vskip 0.3in
]



\printAffiliationsAndNotice{}  

\begin{abstract}
In this paper, we propose a novel Hadamard Transform (HT)-based neural network layer for hybrid quantum-classical computing. It implements the regular convolutional layers in the Hadamard transform domain. The idea is based on the HT convolution theorem which states that the dyadic convolution between two vectors is equivalent to the element-wise multiplication of their HT representation. Computing the HT is simply the application of a Hadamard gate to each qubit individually, so the HT computations of our proposed layer can be implemented on a quantum computer. Compared to the regular Conv2D layer, the proposed HT-perceptron layer is computationally more efficient.  Compared to a CNN with the same number of trainable parameters and 99.26\% test accuracy, our HT network reaches 99.31\% test accuracy with 57.1\% MACs reduced in the MNIST dataset; and in our ImageNet-1K experiments, our HT-based ResNet-50 exceeds the accuracy of the baseline ResNet-50 by 0.59\% center-crop top-1 accuracy using 11.5\% fewer parameters with 12.6\% fewer MACs. 
\end{abstract}

\section{Introduction}
Recently, with the rapid progress in quantum computing hardware, implementing deep neural networks on a quantum computer such as IBM Quantum System (IBM-Q) become feasible for researchers~\cite{matsuo2019reducing,koppenhofer2020quantum}. On the microscopic scale, a quantum is indivisible. All well-known microscopic particles such as electrons and photons are manifestations of the quantum. Classical computers use 0 and 1 to store and process data, while quantum computers' basic computing unit, which is called the qubit, can be both 0 and 1 at the same time. It allows the co-existence of ``superposition states" and thus has more powerful parallel capabilities. Therefore, a quantum computer could perform certain calculations exponentially faster than any modern classical computer. 

However, quantum neural networks (QNN) use as many qubits as the size of the input, which makes it unlikely to be implemented on current quantum computers to solve real-world problems. Therefore, researchers investigate hybrid quantum-classical neural networks instead of pure quantum domain computing. 

Fourier convolution theorem states that the convolution of two vectors in the one-dimensional (1D) time domain and the two-dimensional (2D) spatial domain is equivalent to the element-wise multiplication in the Fourier domain. However, Discrete Fourier Transform (DFT) is a complex transform and it is difficult to implement it using quantum computing. On the other hand, Hadamard Transform (HT) is a binary transform and also holds a similar convolution transform. The Hadamard convolution theorem, which will be reviewed in Section~\ref{sec: Hadamard Convolution Theorem}, inspires us to design a hybrid quantum-classical implementable convolutional layer for deep neural networks. The HT of a vector can be computed in $O(1)$ time using the quantum Hadamard gates. In this paper, we use a hybrid quantum-classical method that takes $O(N)$ time to compute the HT for a $N$-length vector, while the classical fast HT algorithm takes $O(N\log_2N)$ time.

In this work, we propose a hybrid quantum-classical neural network layer based on the HT, and the proposed layer can be used to replace the classical Conv2D layer. It reduces the computational cost significantly by producing comparable or better accuracy results than the baseline classical convolutional model. Our proposed layer is trained as the classical neural network layers using the back-propagation algorithm as it can also be implemented in a purely classical manner. 

Related works include the following:

{\bf WHT-Based Neural Networks} 
Walsh-Hadamard Transform (WHT) is the permuted version of the HT. In~\cite{zhao2021zero}, authors use WHT is used to assist their "ZerO Initialization" method. They apply the WHT in the skip connections, but they did not use the Hadamard convolution theorem and they did not replace the convolutional layers with multiplicative layers in the Hadamard domain. The method described in~\cite{zhao2021zero} does not reduce the number of parameters or the computational cost. Their goal is to improve the accuracy of their networks. Other WHT-based neural networks including~\cite{deveci2018energy, pan2021fast, park2022energy,pan2022block} are used to reduce the computational cost. In~\cite{deveci2018energy}, a binary neural network with a two-stream structure is proposed, where one input is the regular image and the other is the WHT of the image, and convolutions are not implemented in the transform domain. The WHT is only applied at the beginning in~\cite{deveci2018energy}. In~\cite{park2022energy}, authors do not take advantage of the Hadamard convolution theorem. They compute the WHT or related binary transforms and apply the convolution in the transform domain. The WHT-based layers in~\cite{pan2021fast,pan2022block} also use take the element-wise multiplication in the WHT domain like this work, but they do not extract any channel-wise features. They only extract the features width-wise and height-wise, as they have a similar scaling layer to this work. On the contrary, we have channel-wise processing layers in the Hadamard domain to extract the channel-wise features and introduce the quantum-computer-based implementation of the network.

{\bf Trainable Soft-Thresholding} The soft-thresholding function is commonly used in wavelet transform domain denoising~\cite{donoho1995noising} and as a proximal operator for the $\ell_1$ norm based optimization problems~\cite{karakucs2020simulation}. With trainable threshold parameters, soft-thresholding and its variants can be employed to remove the noise in the transform domain-based networks~\cite{badawi2021discrete,pan2021fast,pan2022block,pan2022deep,pan2023real}. 

{\bf QNN, QCNN, and QCCNN} Quantum neural networks (QNNs) based on the principles of quantum mechanics were proposed in~\cite{kak1995quantum}, and quantum convolutional neural networks (QCNNs) were proposed in~\cite{cong2019quantum}. The training method for the QNNs was proposed in~\cite{ricks2003training}. A QCNN uses as many qubits as the size of the input, which makes it unlikely to be implemented on current quantum computers to solve real-world problems. To combine classical neural networks with the advantages of quantum information in order to develop more efficient algorithms, the hybrid quantum-classical convolutional neural networks (QCCNNs) were proposed in~\cite{liu2021hybrid}. To create a QCCNN, the hidden layers can be implemented using parameterized quantum circuits, where the rotation angles for each quantum gate are specified by the components of a classical input vector. The outputs from the previous layer are collected and used as the inputs for the parameterized circuit, and the outputs of the quantum circuits can be obtained from the measurement statistics.

{\bf Hybrid Quantum-Classical WHT} Hybrid quantum-classical algorithms for the Walsh-Hadamard Transform (WHT) are proposed in~\cite{shukla2023hybrid, shukla2023hybrid2}, where the hybrid WHT is used to obtain the WHT in the classical domain for the digital images in the classical domain, but using the Quantum Hadamard gate for computational efficiency. Other procedures are still operated using classical methods. In~\cite{shukla2023hybrid}, authors propose an image denoising method by only changing the high-frequency components in the transform domain to 0. This change is implemented in a classical manner. Then, they apply the hybrid inverse WHT to obtain denoised images, which are essentially low-pass filtered versions of the original images. In~\cite{shukla2023hybrid2}, authors use the hybrid WHT to solve differential equations. The above two papers are not related to neural networks. 

{\bf Other Transform-Based Neural Networks}  The Discrete Fourier Transform (DFT) and Discrete Cosine Transforms are the two of the most important signal and image processing tools. One-dimensional (1D) and two-dimensional (2D) convolutions can be implemented using element-wise multiplications in the 1D and 2D transform domains, respectively. DFT-based neural networks include~\cite{chi2020fast, mohammad2021substitution}, and DCT-based neural networks include~\cite{gueguen2018faster, dos2020good, dos2021less, xu2021dct, ulicny2022harmonic}. However, the DFT is a complex transformation, and Quantum Fourier Transform (QFT) is actually implemented using the Hadamard gates~\cite{weinstein2001implementation}. On the other hand, the HT is implemented only using additions and subtractions and it can be implemented very efficiently using quantum computers. 
Therefore, the HT is more efficient than other transforms.  

\section{Methodology}
\subsection{Background: Hybrid quantum-classical approach for Hadamard Transform}
The HT is a member of generalized Fourier transforms. It can be considered as constructed from size-2 DFT's. As a result, the transform matrix only has  $\pm1$ instead of complex exponential weights as in DFT. Let $\mathbf{x}=[x_0\   x_1\ \ldots\ x_{N-1}]^T$ be a vector with $N=2^M$ components for $M\in\mathbb{N}$, its HT vector
$\mathbf{X} = [X_0\   X_1\ \ldots\ X_{N-1}]^T$ is computed as
\begin{equation}
    X_k = \sqrt\frac{1}{N} \sum_{m=0}^{N-1}x_m(-1)^{\sum_{i=0}^{M-1}k_im_i},
\end{equation}
where, $k_i$ and $m_i$ are the $i$-bits in the binary representations of $k$ and $m$, respectively. The HT can be computed as the matrix product between $\mathbf{X}$ and the Hadamard matrix $\mathbf{H}_N$:
\begin{equation}
    \mathbf{X}=\mathcal{H}(\mathbf{x})=\sqrt\frac{1}{N} \mathbf{H}_N\mathbf{x}.
\end{equation}

The Hadamard matrix $\mathbf{H}_N$ is constructed as
\begin{equation}\label{eq: Hadamard matrix}
			\mathbf{H}_N = 
			\begin{cases}
				1,& N = 1,\\
				\begin{bmatrix}
					1 & 1 \\ 1 & -1
				\end{bmatrix},& N =2,\\
				\begin{bmatrix}
					\mathbf{H}_{\frac{N}{2}} & \mathbf{H}_{\frac{N}{2}} \\ \mathbf{H}_{\frac{N}{2}} & -\mathbf{H}_{\frac{N}{2}}
				\end{bmatrix},& N \ge 4,
			\end{cases}
		\end{equation}
		Alternatively, for $N\ge4$, $N=2^M$, $\mathbf{H}_N$ can also be computed using Kronecker product $\otimes$:
		\begin{equation}
			\mathbf{H}_{N}=\mathbf{H}_2 \otimes\mathbf{H}_{\frac{N}{2}}=\mathbf{H}_2^{\otimes M}.
		\end{equation}

The Walsh matrix $\mathbf{W}_N$ for the Walsh-Hadamard Transform (WHT) is the sequency-order-rearranged version of the Hadamard matrix $\mathbf{H}_N$~\cite{walsh1923closed, yuen1972remarks}. The sequency ordering can be derived from the natural ordering (Hadamard ordering) by first applying the bit-reversal permutation and then the Gray-code permutation. 

  With the squency ordering, the number of sign changes in a row is in increasing order. The more changes there are, the higher the frequency component is extracted. Therefore, the Walsh matrix is more commonly used. However, in this work, we apply element-wise multiplication in the Hadamard domain. Therefore, permutation is unnecessary in this work, so we use HT instead of WHT to simplify the implementation.

The Hadamard matrix $\mathbf{H}_N$ is orthogonal and symmetric, as 
\begin{equation}
    \mathbf{H}_N\mathbf{H}_N^T=N\mathbf{I}_N, \mathbf{H}_N=\mathbf{H}_N^T, 
\end{equation}
so the Inverse Hadamard Transform (IHT) can be implemented in a similar manner as the forward HT:
\begin{equation}
    \mathbf{x}=\mathcal{H}^{-1}(\mathbf{X})=\sqrt\frac{1}{N} \mathbf{H}_N\mathbf{X}=\mathcal{H}(\mathbf{X}).
\end{equation} 
In practice, we can combine two $\sqrt\frac{1}{N}$ normalization terms from the HT and the IHT to one $\frac{1}{N}$ to avoid the square-root operation.

Similar to the Fast Fourier Transform (FFT), HT can be implemented in a fast way using the butterfly structures as Eq. (1) in~\cite{fino1976unified}. In this method, the Fast Hadamard Transform (FHT) has the time complexity of $O(N\log_2N)$ in the classical domain. 

On the other hand, $\mathbf{H}=\sqrt\frac{1}{2}\mathbf{H}_2=\sqrt\frac{1}{2}\begin{bmatrix}1& 1 \\ 1 & -1\end{bmatrix}$ is the transformation matrix of the quantum Hadamard gate in a computational bias. Therefore, the Hadamard transform can be computed in $O(1)$ time in the quantum domain, as it is a quantum logic gate that can be parallelized. Let $\mathbf{\bar{x}}=[\bar{x}_0\   \bar{x}_1\ \ldots\ \bar{x}_{N-1}]^T$ be a normalized version of $\mathbf{x}$, i.\,e, $\mathbf{\bar{x}}= \frac{1}{ ||\mathbf{x}||}\mathbf{x}$, the quantum implementation of HT of $\mathbf{\bar{x}}$ involves preparing the initial state $\sum_{k=0}^{N-1}\bar{x}_k|k\rangle$, and then applying quantum Hadamard gates $\mathbf{H}^{\otimes M}$ on it. It can be verified that
\begin{equation}
\begin{split}
\mathbf{H}^{\otimes M}\left[\sum_{k=0}^{N-1}\bar{x}_k|k\rangle\right] &= \sqrt\frac{1}{N} \sum_{k=0}^{N-1}\sum_{m=0}^{N-1}\bar{x}_m(-1)^{\sum_{i=0}^{M-1}k_im_i} \\&=\sum_{k=0}^{N-1}\bar{X}_k|k\rangle,
\end{split}    
\end{equation}
where, $\mathbf{\bar{X}}$ is the HT of $\mathbf{\bar{x}}$, $N=2^M$. Although the quantum approach for the HT on a $M$-length vector has the computational cost of $O(1)$, the difficulty lies in the measurement. One can only find the square of the amplitude of the HT values by carrying out the measurement. However, the HT of $\mathbf{\bar{x}}$ has both positive and negative values, but the sign information is lost during the amplitude measurement. To obtain the correct sign information for the HT, we adopt the Algorithm 1 in~\cite{shukla2023hybrid2}, which is based on Lemma~\ref{lemma: HT positive} and increases the computational cost to $O(N)$. 

\begin{lemma}\label{lemma: HT positive}
Let $\mathbf{x}=[x_0\   x_1\ \ldots\ x_{N-1}]^T$ and $\mathbf{X} = \mathcal{H}(\mathbf{x})=[X_0\   X_1\ \ldots\ X_{N-1}]^T$. If $x_0 > \sum_{k=1}^{N-1}|x_k|$, then $X_k>0$ for $k=0, 1, ..., N-1$.
\end{lemma}

The proof for Lemma~\ref{lemma: HT positive} is presented in Appendix~\ref{proof: HT positive}. We can change $x_0$ to a large number to satisfy the condition in Lemma~\ref{lemma: HT positive}. The HT of the revised vector will have all positive entries that can be computed in the quantum domain efficiently. After that, we can revise the HT results based on the change on $x_0$ to get the correct $\mathbf{X}=\mathcal{H}(\mathbf{x})$.

\begin{algorithm}[tb]
   \caption{The hybrid quantum-classical HT algorithm.}
   \label{alg: QHT}
\begin{algorithmic}
   \STATE {\bfseries Input:} The input vector $\mathbf{x}= [x_0\   x_1\ \ldots\ x_{N-1}]^T\in\mathbb{R}^N$ where $N=2^M$, $M\in\mathbb{N}$.
   \STATE {\bfseries Output:} The output vector $\mathbf{X}= [x_0\   x_1\ \ldots\  x_{N-1}]^T$ which is the HT of the $\mathbf{x}$.
   \STATE $b=\epsilon+\sum_{k=0}^{N-1}|x_k|$, where $\epsilon$ is any positive number;
   \STATE $\mathbf{\tilde{x}}= [b\  x_1\   x_2\ \ldots\ x_{N-1}]^T$;
   \STATE $c=||\mathbf{\tilde{x}}||=\sqrt{b^2+\sum_{k=1}^{N-1} x_k^2}$.
   \STATE $\mathbf{\bar{x}}= [\tilde{x}_0\   \tilde{x}_1\ \ldots\ \tilde{x}_{N-1}]^T = \mathbf{\tilde{x}}/c$;
   \STATE Prepare the state $|\Psi\rangle=\sum_{k=0}^{N-1}\bar{x}_k|k\rangle$ using $M$ qubits;
   \STATE Apply $\mathbf{H}^{\otimes M}$ on $|\Psi\rangle$;
   \STATE Measure all the $M$ qubits to compute the probability $p_k$ of obtaining the state $|k\rangle$ for $k=0$ to $N-1$;
   \STATE $\delta = \sqrt\frac{1}{N}(b-\mathbf{x}[0])$;
   \STATE $\mathbf{X}=[c\sqrt{p_0}-\delta\ c\sqrt{p_1}-\delta\ \ldots\  c\sqrt{p_{N-1}}-\delta]^T$.
\end{algorithmic}
\end{algorithm}

We use Algorithm 1 in~\cite{shukla2023hybrid2} to implement the HT in our deep neural network. In~\cite{shukla2023hybrid2} they used WHT in Algorithm~\ref{alg: QHT}, instead, we use HT. In summary, we first change the first element of the vector $x_0$ to a large number $b$ and reconstruct the vector as $\mathbf{\tilde{x}}$, then we normalize it by $c=||\mathbf{\tilde{x}}||$ as $\mathbf{\bar{x}}$. We apply such a normalization because the norm of the input vector to the quantum state must be 1. After that, we prepare the state $|\Psi\rangle=\sum_{k=0}^{N-1}\bar{x}_k|k\rangle$ and apply quantum Hadamard gates $\mathbf{H}^{\otimes M}$ on $|\Psi\rangle$. Then we measure all the $M$ qubits to compute the probability $p_k$ of obtaining the state $|k\rangle$ for $k=0$ to $N-1$. These $p_k$ are the scaled version of the square amplitude of the HT of $\mathbf{\tilde{x}}$, so we scale the $\sqrt{p_k}$ by the normalization factor $c$. Finally, we subtract the results by $\delta = \sqrt\frac{1}{N}(b-x_0)$ to obtain the HT of $\mathbf{x}$.

In this work, we design the proposed neural network layer based on the Two-Dimensional Hadamard transform (HT2D). Let $\mathbf{x}=\begin{bmatrix}
  x_{0,0} &\cdots& x_{0,N-1}\\ 
  \vdots &\ddots& \vdots\\ 
  x_{N-1,0} &\cdots& x_{N-1,N-1}\\ 
\end{bmatrix}$, $N=2^M$, $M\in\mathbb{N}$. The HT expression $\mathbf{X}=\begin{bmatrix}
  X_{0,0} &\cdots& X_{0,N-1}\\ 
  \vdots &\ddots& \vdots\\ 
  X_{N-1,0} &\cdots& X_{N-1,N-1}\\ 
\end{bmatrix}$ is
\begin{equation}
\begin{split}
    & X_{k,l} =\frac{1}{N} \sum_{m=0}^{N-1}\sum_{n=0}^{N-1}x_{m,n}(-1)^{\sum_{i=0}^{M-1}k_im_i+l_in_i}\\
    & =\frac{1}{N} \sum_{m=0}^{N-1}\left(\sum_{n=0}^{N-1}x_{m,n}(-1)^{\sum_{i=0}^{M-1}l_in_i}\right)(-1)^{\sum_{i=0}^{M-1}k_im_i},\\
\end{split}
\end{equation}
where, $k_i$, $l_i$, $m_i$ and $n_i$ are the $i$-bits in the binary representations of $k$, $l$, $m$ and $n$. Therefore, the HT2D can be obtained from the one-dimensional Hadamard transform (HT1D) in a separable manner for computational efficiency~\cite{vetterli1985fast}, as Algorithm~\ref{alg: QHT2D} presents. The complexity of HT2D on an $N\times N$ image is $O(N^2)$ (where $N$ is an integer power of 2) using Algorithm~\ref{alg: QHT2D}. On the other hand, the complexity of the classical two-dimensional Fast Hadamard transform (FHT2D) is $O(N^2\log_2 N)$. 
\begin{algorithm}[tb]
   \caption{The two-dimensional HT algorithm. $\mathbf{X}[i]$  and $\mathbf{X}^T[i]$ represent the $i$-th column and row of $\mathbf{X}$.}
   \label{alg: QHT2D}
\begin{algorithmic}
   \STATE {\bfseries Input:} The input matrix $\mathbf{x}\in\mathbb{R}^{N_1\times N_2}$.
   \STATE {\bfseries Output:} The output vector $\mathbf{X}\in\mathbb{R}^{N_1\times N_2}$ which is the HT of the $\mathbf{x}$.
   \FOR{$i=0$ {\bfseries to} $N_1-1$}
   \STATE $\mathbf{X}[i]=\mathcal{H}(\mathbf{x}[i])$ using Algorithm~\ref{alg: QHT};
   \ENDFOR
    \FOR{$i=0$ {\bfseries to} $N_1-1$}
   \STATE $\mathbf{X}^T[i]=\mathcal{H}(\mathbf{X}^T[i])$ using Algorithm~\ref{alg: QHT};
   \ENDFOR
\end{algorithmic}
\end{algorithm}

\subsection{Hadamard Convolution Theorem}~\label{sec: Hadamard Convolution Theorem}
Fourier convolution theorem states that an input feature map $\mathbf{x}\in\mathbb{R}^N$ and a kernel $\mathbf{a}\in\mathbb{R}^K$ can be convolved in the Fourier domain as follows:
\begin{equation}
\mathbf{a} *_c\mathbf{x} = \mathcal{F}^{-1}\left(\mathbf{A}\circ\mathbf{X}\right), 
\end{equation}
where, $\mathcal{F}(\cdot)$ stands for DFT and $\mathcal{F}^{-1}(\cdot)$ stands for IDFT. $\mathbf{A}=\mathcal{F}(\mathbf{a})$, $\mathbf{X}=\mathcal{F}(\mathbf{x})$. $*_c$ is the circular convolution operator and $\circ$ represents the element-wise multiplication. Similar to the Fourier convolution theorem, the Hadamard convolution theorem holds as Theorem~\ref{Hadamard convolution theorem}, which states the dyadic convolution in the time domain is equivalent to the element-wise multiplication in the Hadamard domain. Theorem~\ref{Hadamard convolution theorem} inspires us to design the HT-based layer which will be discussed in the following section to replace the convolutional layers in convolutional neural networks (CNNs).
\begin{theorem}[Hadamard convolution theorem]\label{Hadamard convolution theorem}
Let $M\in\mathbb{N}$, $N=2^M$, and $\mathbf{a}, \mathbf{x}\in\mathbb{R}^N$. The convolution $ \mathbf{y}=\mathbf{a}*_d\mathbf{x}\Longleftrightarrow$ the element-wise multiplication  $Y_k = A_kX_k$ for $k=0, 1, \ldots, N-1$, where, $\mathbf{Y}=\mathcal{H}(\mathbf{y})=[Y_0\ Y_1\ \ldots\ Y_{N-1}]^T$, $\mathbf{A}=\mathcal{H}(\mathbf{a})=[A_0\ A_1\ \ldots\ A_{N-1}]^T$, and $\mathbf{X}=\mathcal{H}(\mathbf{x})=[X_0\ X_1\ \ldots\ X_{N-1}]^T$. 
\end{theorem}

The proof of Theorem~\ref{Hadamard convolution theorem} is presented in Appendix~\ref{proof: Hadamard convolution theorem}~\cite{gulamhusein1973simple,uvsakova2002using, uvsakova2002walsh, gajic2011calculation}. Although the dyadic convolution $*_d$ is not the same as circular convolution, we will use the HT for convolutional filtering in neural networks. This is because 
HT is also related to the block Haar wavelet packet transform~\cite{cetin1993block} and each Hadamard coefficient approximately represents a frequency band. As a result, applying weights onto frequency bands and computing the inverse HT is an approximate way of frequency domain filtering similar to the Fourier transform-based convolutional filtering.


\subsection{HT-Perceptron Layer}
\begin{figure}[tb]
\vskip 0.2in
\begin{center}
\centerline{\includegraphics[width=1\linewidth]{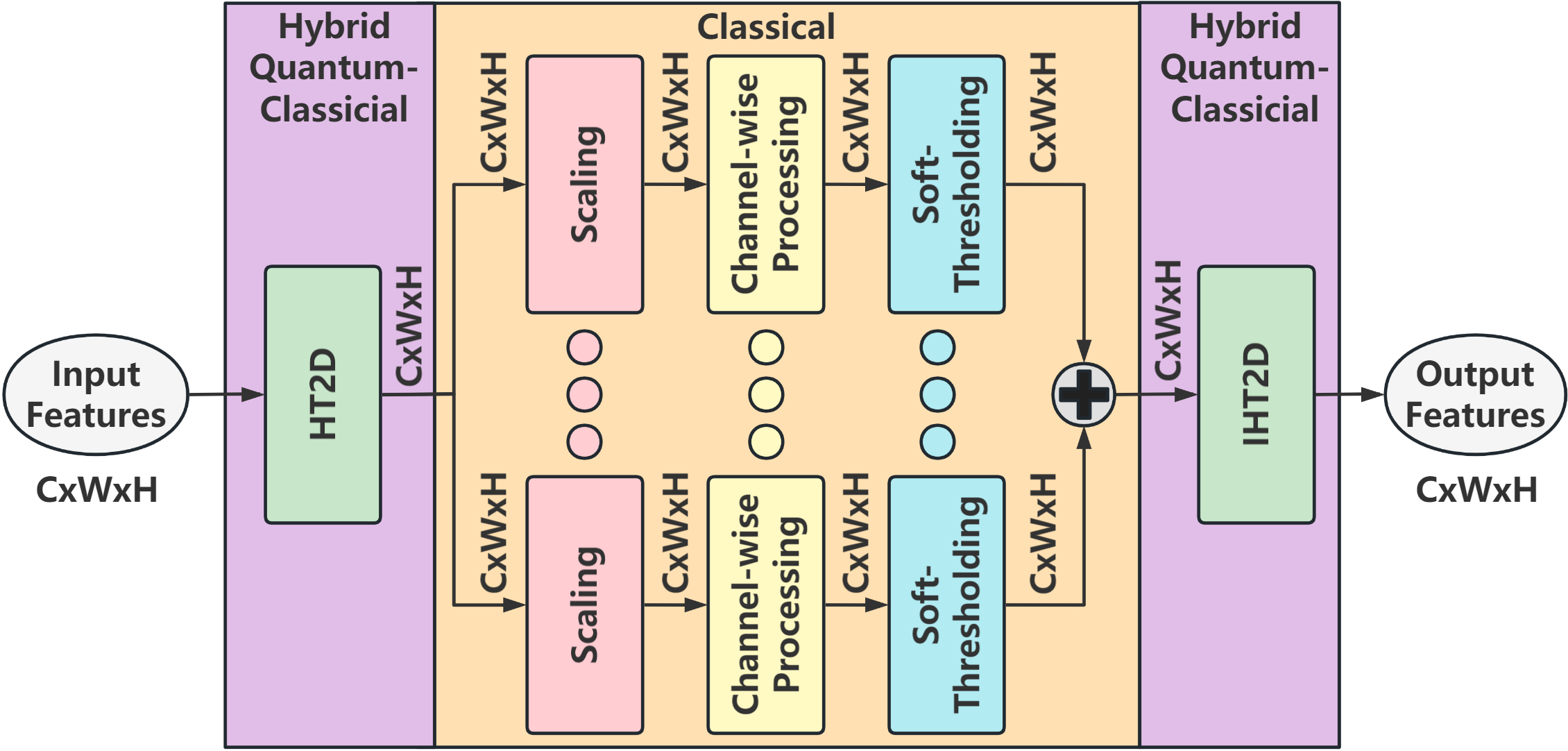}}
\caption{Structure of the proposed HT-perceptron layer for a tensor in $\mathbb{R}^{C\times W\times H}$. The HT2D and IHT2D are implemented using  the quantum computer~\ref{alg: QHT2D} (or the classical fast approach) and multiplications and soft-thresholding operations of the network are implemented using the classical approach. We have parallel multiple paths to increase the number of trainable parameters. Each path corresponds to a convolutional kernel. If we want to change the number of output channels, we can change the number of kernels at each channel-wise processing.}
\label{fig: ht-percpetron}
\end{center}
\vskip -0.2in
\end{figure}

The proposed HT-perceptron layer is presented in Figure~\ref{fig: ht-percpetron} and Algorithm~\ref{alg: HT-Perceptron Layer}. The HT-perceptron layers are combined to construct deep ``convolutional" neural networks as shown in Tables~\ref{tab: resnet-20} and~\ref{tab: resnet-50} in Appendix~\ref{Figures and Tables}. They replace the Conv2D layers in a typical CNN, therefore, we apply an HT2D along the width and height of the tensor. Similar to a Conv2D layer which contains multiple kernels, our structure has multiple parallel paths. In each path, we first apply element-wise multiplications on the tensor with a $W$ by $H$ trainable matrix $\mathbf{A}$. This operation is equivalent to convolutional filtering and we call it scaling because the HT coefficients are scaled as in Fourier domain filtering. In this work, we initialize $\mathbf{A}$ as random numbers from the uniform distribution of $[0, 1)$. Then, we perform channel-wise processing. This step is implemented similarly to the so-called $1\times 1$ convolution of the Conv2D layer. If we want to change the number of output channels, we can change the number of kernels at this step. After scaling and $1\times 1$ convolution we apply a soft-thresholding function as a nonlinearity instead of RELU because both positive and negative amplitudes are important in the transform domain. Soft-thresholding is widely used in wavelet denoising to remove noise from the data~\cite{chang2000adaptive,zhao2015improved}. Parameters of the soft-thresholding nonlinearity can be trainable, i.\,e, they can be learned using the back-propagation algorithm. Finally, we apply an IHT2D on the summation of all paths to obtain the resulting tensor output of the HT-perceptron layer.

\begin{figure}[tb]
\vskip 0.2in
\begin{center}
\centerline{\includegraphics[width=1\linewidth]{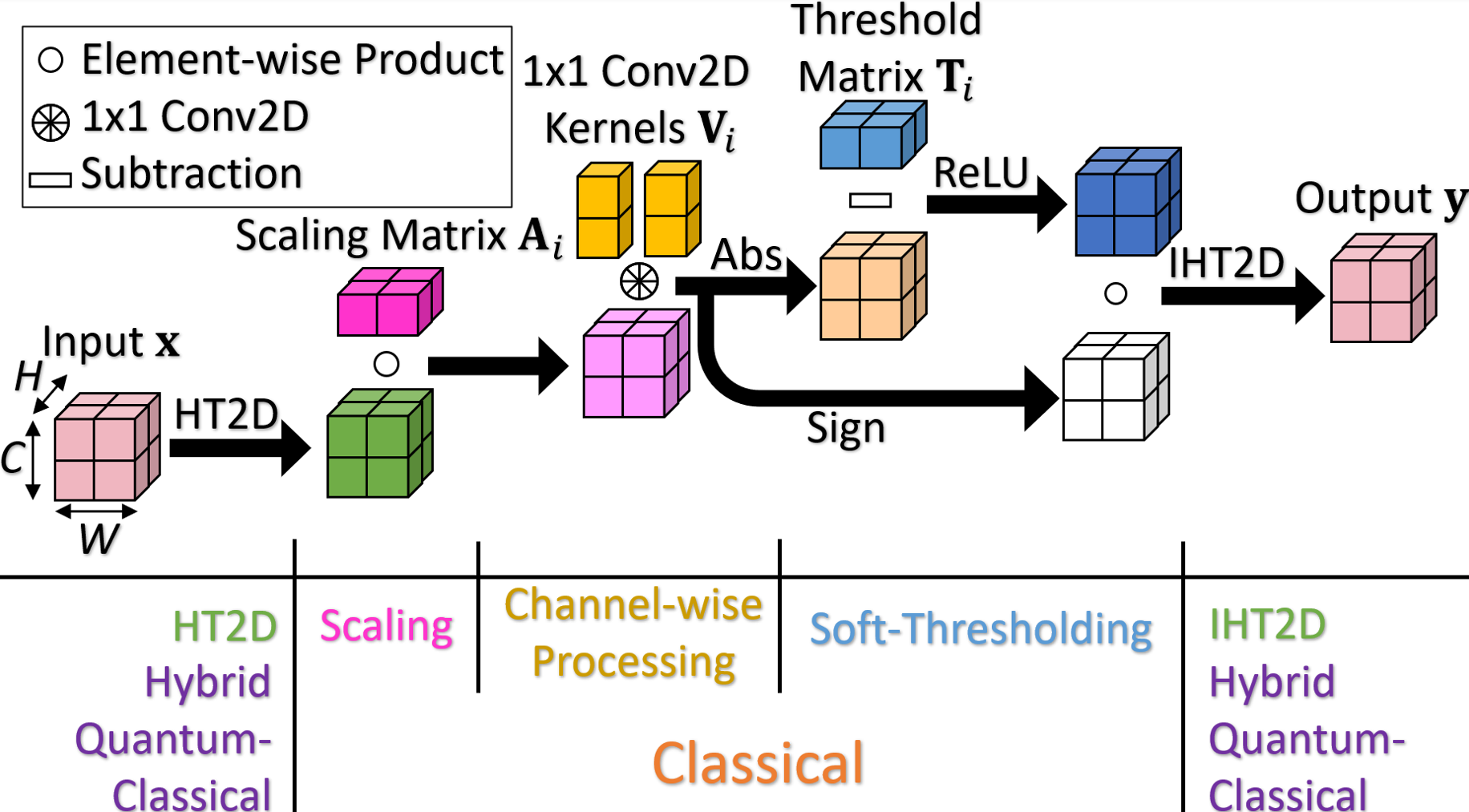}}
\caption{Procedure of each path in the HT-Perceptron layer. Each entry along the width and the height is processed individually in the Hadamard domain, so we don't need to apply the permutation as the Walsh-Hadamard transform.}
\label{fig: path}
\end{center}
\vskip -0.2in
\end{figure}

\begin{algorithm}[tb]
   \caption{$P$-Path HT-Perceptron Layer. Dimensions are defined as batch size, channel, height, and width.}
   \label{alg: HT-Perceptron Layer}
\begin{algorithmic}
   \STATE {\bfseries Input:} The input tensor $\mathbf{x}\in\mathbb{R}^{B\times C_i\times H\times W}$.
   \STATE {\bfseries Output:} The output tensor $\mathbf{y}\in\mathbb{R}^{B\times C_o\times H\times W}$.
   \STATE {\bfseries Define:} $\mathbf{A}_i, \mathbf{T}_i\in\mathbb{R}^{H\times W}$, $\mathbf{V}_i=$ Conv2D(in channels$=C_i$, out channels$=C_o$, kernel size$=1$), {\bfseries for} $i=0$ {\bfseries to} $P$.
   \STATE $\mathbf{X}=\text{HT2D}(\mathbf{x})$;
   \FOR{$i=0$ {\bfseries to} $P-1$}
   \STATE $\mathbf{X}_i=\mathbf{X}\circ\mathbf{A}_i$;
   \STATE $\mathbf{Z}_i=\mathbf{V}_i(\mathbf{X}_i)$;
   \STATE $\mathbf{Y}_i=\text{sign}(\mathbf{Z}_i)\circ\text{ReLU}(|\mathbf{Z}_i|-\mathbf{T}_i)$;
   \ENDFOR
   \STATE $\mathbf{Y}=\text{sum}(\text{stack}(\mathbf{Y}_i))$;
   \STATE $\mathbf{y}=\text{IHT2D}(\mathbf{Y})$.
\end{algorithmic}
\end{algorithm}

Trainable soft-thresholding is applied to remove small entries in the Hadamard domain. This operation is similar to image coding and transform-domain denoising~\cite{wallace1991jpeg, le1991mpeg}. It is defined as follows:
\begin{equation}
    \mathbf{Y} = \mathcal{S}_\mathbf{T}(\mathbf{X}) = \text{sign}(\mathbf{X})\circ(|\mathbf{X}|-\mathbf{T})_+,
\end{equation}
where, $\circ$ stands for the element-wise multiplication, $(\cdot)_+$ stands for the ReLU function, and $\mathbf{T}$ stands for non-negative trainable threshold parameters which are determined using the back-propagation algorithm. $\mathbf{T}\in\mathbb{R}^{W \times H}$ if $\mathbf{X}\in\mathbb{R}^{C\times W \times H}$. In this work, we initialize $\mathbf{T}$ as random numbers from the uniform distribution of $[0, 0.1)$. We initialize $\mathbf{T}$ with small positive values because it is the threshold parameter to remove small valued feature maps which correspond to noise. The ReLU function is not suitable because both significantly large positive and negative values are important in the Hadamard domain. For example, a completely positive vector can have both significant positive and negative values in the Hadamard domain. Furthermore, the multiplication between $\text{sign}(\mathbf{X})$ and $(|\mathbf{X}|-\mathbf{T})_+$ can be implemented using sign-bit operations, so no multiplication operation is required for the soft-thresholding.

In summary, an HT-perceptron layer with $P$ paths mapping from $\mathbf{x}$ to $\mathbf{y}$ is defined as:
\begin{equation}    
\mathbf{y}=\mathcal{H}^{-1}\left(\sum_{i=0}^{P-1}\mathcal{S}_{\mathbf{T}_i}\left(\mathcal{H}(\mathbf{x})\circ\mathbf{A}_i\circledast\mathbf{V}_i\right)\right),
\end{equation}
where, $\mathcal{H}(\cdot)$ and $\mathcal{W}^{-1}(\cdot)$ stand for HT2D and IHT2D, $\mathbf{A}_i$ is the scaling matrix in the $i$-th path, $\mathbf{V}_i$ represents the $1\times 1$ Conv2D kernels used in the channel-wise processing in the $i$-th path, $\mathbf{T}_i$ is the threshold parameter matrix in the soft-thresholding layer in the $i$-th path, $\circ$ stands for the element-wise multiplication, and $\circledast$ represents channel-wise processing which can be implemented using the $1\times 1$ Conv2D layer. The procedure in each path of the HT-Perceptron layer is illustrated in Figure~\ref{fig: path}.

Table~\ref{tab: parameters} shows the number of parameters in an HT-perceptron layer is $2PN^2+PC^2$. The input and output tensors are assumed to be in $\mathbb{R}^{C\times N\times N}$. In each path, there are $N^2$ parameters in the scaling matrix, $N^2$ parameters in the soft-threshold matrix, and $C^2$ parameters in the channel-wise $1\times 1$ convolution. Therefore, a $P$-path HT-perceptron layer has $2PN^2+PC^2$ parameters. For comparison, a $K\times K$ Conv2D layer has $KC^2$ parameters. A 3-path HT-perceptron layer has the same amount of parameters as a $3\times 3$ Conv2D layer if $C=N$. Furthermore, in the most main-stream CNNs such as ResNets~\cite{he2016deep}, $C$ is usually much larger than $N$ in the hidden layers, then our proposed HT-perceptron layer can save parameters for ResNet-type CNNs as discussed in Section~\ref{sec: Experimental Results}.

\begin{table}[tb]
\begin{center}
\begin{small}
\begin{sc}
    \caption{Parameters of a Conv2D Layer Versus an HT-perceptron layer for a $C$-channel $N\times N$ image. $N$ is an integer power of 2.}
    \label{tab: parameters}
\vskip 0.15in
    \centering
    \begin{tabular}{lc}
    \toprule
        Layer (Operation)&Parameters\\
        \midrule
         $K\times K$ Conv2D& $K^2C^2$\\
         $3\times 3$ Conv2D& $9C^2$\\
        \midrule
         HT2D&$0$\\
         Scaling& $N^2$\\
         Channel-wise Processing & $C^2$\\
         Soft-Thresholding& $N^2$\\
         IHT2D& $0$\\
        \midrule
         $P$-path HT-perceptron & $2PN^2+PC^2$\\
         $1$-path HT-perceptron & $2N^2+C^2$\\
         $3$-path HT-perceptron & $6N^2+3C^2$\\
         $5$-path HT-perceptron & $10N^2+5C^2$\\
        \bottomrule
         \end{tabular}
\end{sc}
\end{small}
\end{center}
\vskip -0.1in
\end{table}

\begin{table}[tb]
\begin{center}
\begin{small}
\begin{sc}
    \caption{Multiply–Accumulates (MACs) of a Conv2D Layer Versus an HT-perceptron layer for a $C$-channel $N\times N$ image. $N$ is an integer power of 2. MACs from the HT2D and the IHT2D are omitted.} 
    \label{tab: MACs}
\vskip 0.15in
    \centering
    \begin{tabular}{lc}
    \toprule
         Layer (Operation)&MACs\\
        \midrule
         $K\times K$ Conv2D&$K^2N^2C^2$\\
         $3\times 3$ Conv2D&$9N^2C^2$\\
        \midrule
         Scaling, soft-thresholding&$N^2C$\\
         Channel-wise processing & $N^2C^2$\\
        \midrule
         $P$-path HT-perceptron & $PN^2C+PN^2C^2$\\
         $1$-path HT-perceptron & $N^2C+N^2C^2$\\
         $3$-path HT-perceptron & $3N^2C+3N^2C^2$\\
         $5$-path HT-perceptron & $5N^2C+5N^2C^2$\\
        \bottomrule
         \end{tabular}
\end{sc}
\end{small}
\end{center}
\vskip -0.1in
\end{table}

If the input tensor and the output tensor are in $\mathbb{R}^{C\times N\times N}$, the computational cost of a $K\times K$ Conv2D is $O(K^2N^2C^2)$. On the other hand, the classical FHT algorithm takes $O(N^2\log_2N)$, and
the hybrid quantum-classical HT algorithm reduces it to $O(N^2)$. Compared to the $3\times 3$ Conv2D whose complexity is $O(N^2C^2)$, the $O(N^2C)$ from the HT and the IHT can be omitted. In each path, the complexity to compute the scaling and the soft-thresholding is $O(N^2C)$, and the complexity of the channel-wise processing is $O(N^2C^2)$. Therefore, the total complexity of a $P$-path HT-perceptron layer is $O(PN^2C^2)$. To compare the total computational cost between the HT-perceptron layer with the Conv2D layer, we use Multiply–Accumulate (MACs). $1$ MAC contains $1$ addition and $1$ multiplication. As Table~\ref{tab: MACs} states, in each path, there are $N^2C$ multiplications in scaling and $N^2C$ additions in the soft-thresholding. There is no addition in the scaling and no multiplication in the soft-thresholding. Thus, we totally need $N^2C$ MACs to compute the scaling and the soft-thresholding in each path. In each path, the channel-wise processing has $N^2C^2$ MACs because the channel-wise processing is implemented using the $1\times 1$ Conv2D layer. Furthermore, MACs from the HT2D and IHT2D can be omitted even when we use the classical FHT approach because the HT2D and the IHT2D can be implemented without any multiplication, as the normalization factor can be computed with the scaling. Therefore, we totally need $PN^2C+PN^2C^2$ MACs to compute a $P$-path HT-perceptron layer. As a comparison, we totally need $K^2N^2C^2$ MACs to compute a $K\times K$ Conv2D layer. Briefly speaking, the proposed HT-perceptron layer reduces some $N^2C^2$ MACs to $N^2C$. In consequence, our proposed HT-perceptron layer is more computationally efficient than the Conv2D layer.

\section{Experimental Results}\label{sec: Experimental Results}
{\bf Hybrid Hadamard Neural Network on MNIST}
We start the experimental section with a toy example of MNIST handwritten digits classification task using a hybrid quantum-classical Hadamard neural network. The MNIST experiments are carried out on the IBM Quantum Lab cloud computer using PyTorch and Qiskit. Since the MNIST image size is $28\times 28$, we pad 2 pixels with 0s on all borders to make the input image size $32\times 32$. We first convert the intensities of the raw images to the  $0$ - $1$ range. We then normalize the MNIST images with a mean of $0.1307$ and a standard deviation of $0.3081$.

\begin{table}[tb]
\begin{center}
\begin{small}
\begin{sc}
\caption{Toy CNN for the MNIST classification task. We pad all borders 2 pixels with 0s to make the image size $32\times 32$. We revise the layer Conv2 using the proposed HT-perceptron layer. }
\label{tab: toy-cnn}
\vskip 0.15in
\centering
    \begin{tabular}{lcc}
    \toprule
		Layer&Output Shape&Implementation\\
        \midrule
		Input&$1\times32\times32$&-\\
		Conv1&$32\times32\times32$&$3\times 3, 32$\\  
            Conv2&$32\times32\times32$&$3\times 3, 32$\\  
		MaxPool&$32\times16\times16$&$2\times 2$\\
		Flatten&$8192$&-\\
            Dense&$128$&Linear, 128\\
		Output&$10$&Linear, 10\\
        \bottomrule
	\end{tabular}
 \end{sc}
\end{small}
\end{center}
\vskip -0.1in
\end{table}

As Table~\ref{tab: toy-cnn} shows, the CNN is built using two $3\times 3$ Conv2D layers, one average pooling layer, and two linear layers. ReLU is used as the activation function in the two convolution layers and the first linear layer. Dropout with a probability of 0.2 is applied after the second Conv2D layer and the first linear layer. Bias terms are applied in all Conv2D and linear layers. The first Conv2D layer increases the number of channels from $1$ to $32$. We retrain the first Conv2D layer and replace the second Conv2D layer using a 3-path HT-perceptron layer. The 3-path HT structure is used here because the input and output shapes of the second Conv2D layer are both $32\times 32\times 32$, so we keep the number of parameters the same for the CNN and the HT-CNN. There are $1,059,592$ parameters in each neural network ($3^2\times1\times32+32=320$ are in the first Conv2D layer, $3^2\times32^2+32=9,248$ parameters are in the second Conv2D layer, $8192\times 128+128=1,048,704$ parameters are in the first linear layer, and $1,290$ parameters are in the second linear layer).

To train the neural networks on the MNIST dataset, we use the Adadelta optimizer~\cite{zeiler2012adadelta} with an initial learning rate of 1.0. The learning rate decays by 0.7 after each epoch. Models are trained with the mini-batch size of 64 for 14 epochs. During the training, the best models are saved based on the accuracy of the MNIST test dataset, and their accuracy results are reported in Table~\ref{tab: mnist}. After replacing the second Conv2D layer by the 3-path HT-perceptron layer, $9\times 32^4-(3\times 32^4+3\times 32^3)=6.19$ million MACs  are reduced ($57.1\%$), while the accuracy even improves from $99.26\%$ to $99.31\%$. 

\begin{table}[tb]
\begin{center}
\begin{small}
\begin{sc}
\caption{MNIST Experimental Results. There are 1,059,562 parameters in each neural network.}
\label{tab: mnist}
\vskip 0.15in
\centering
    \begin{tabular}{lcc}
    \toprule
	Method&MACs (M)&Accuracy\\
        \midrule
	CNN&10.85&99.26\%\\
        HT-CNN (3-path)&4.66 (57.1\%$\downarrow$)&99.31\%\\
        \bottomrule
	\end{tabular}
 \end{sc}
\end{small}
\end{center}
\vskip -0.1in
\end{table}

{\bf Hadamard Network on CIFAR-10 and ImageNet-1K}
Experiments on the CIFAR-10 and ImageNet-1K are carried out on a workstation computer with an NVIDIA RTX3090 GPU using PyTorch. We don't use the IBM-Q cloud platform 
in these experiments because the cloud platform is too slow for large datasets.
We use the classical Fast-HT algorithm instead of Algorithm~\ref{alg: QHT2D}.  
In these experiments, ResNet-20 for the CIFAR-10 classification task and ResNet-50 for the ImageNet-1K classification task~\cite{he2016deep} are used as the backbone networks.

To revise ResNet-20 and ResNet-50, we retain the first $3\times 3$ Conv2D layer, then we replace those $3\times 3$ Conv2D layers at the even indices (the second Conv2D in each convolutional block in ResNet-20, and Conv2\_2, Conv3\_2, Conv3\_4, etc., in ResNet-50) with the HT-perceptron layer. We keep the $3\times3$ Conv2D layers at odd indices because, in this way, we use the regular $3\times 3$ Conv2D layer and the proposed HT-perceptron layer by turns, then the network can extract features in different manners efficiently. We call those HT-perceptron-layer-revised ResNets as HT-ResNets. Tables~\ref{tab: resnet-20} and~\ref{tab: resnet-50} in Appendix~\ref{Figures and Tables} describe the method we revising the ResNet-20 and ResNet-50.

To train ResNet-20 and the HT-ResNet-20s, We use the SGD optimizer with a weight decay of 0.0001 and momentum of 0.9. Models are trained with a mini-batch size of 128 for 200 epochs. The initial learning rate is 0.1, and the learning rate is reduced by 1/10 at epochs 82, 122, and 163, respectively. Data augmentation is implemented as follows: First, we pad 4 pixels on the training images. Then, we apply random cropping to get 32 by 32 images. Finally, we randomly flip images horizontally. We normalize the images with the means of [0.4914, 0.4822, 0.4465] and the standard variations of [0.2023, 0.1994, 02010]. During the training, the best models are saved based on the accuracy of the CIFAR-10 test dataset, and their accuracy numbers are reported in Table~\ref{tab: CIFAR-10}. 

In the CIFAR-10 experiments, we try different numbers  of paths ranging from 1 to 6 in the HT-ResNet-20s. As shown in Table~\ref{tab: CIFAR-10},
the best accuracy (91.58\%) is obtained from the 5-path structure among the HT-ResNet-20s with 1 to 6 paths. Although in the CIFAR-10 experiments, we do not obtain any better accuracy than our baseline model, we successfully reduce many parameters and MACs with producing comparable accuracy results, as the accuracy of the 5-path HT-ResNet-20 drops less than 0.08\% compared to the baseline ResNet-20. Our HT-ResNet-20s obtains higher accuracy results than~\cite{pan2022block} because we extract the channel-wise features using $1\times 1$ Conv2D in the Hadamard domain.

\begin{table*}[t]
\begin{center}
\begin{small}
\begin{sc}
\caption{CIFAR-10 Experimental Results.}
    \label{tab: CIFAR-10}
    \centering
    \begin{tabular}{lccc}
    \toprule
       Method&Parameters&MACs (M)&Accuracy\\
        \midrule
        ResNet-20 \cite{he2016deep}&0.27M&-&91.25\%\\
        WHT-based ResNet-20~\cite{pan2022block}&133,082 (51.26\%$\downarrow$)&-&90.12\%\\
        \midrule
        ResNet-20 (our trial, baseline)&272,474&41.32&91.66\%\\
        HT-ResNet-20 (1-path)&151,514 (44.39\%$\downarrow$)&22.53 (45.5\%$\downarrow$)&91.25\%\\
        HT-ResNet-20 (2-path)&175,706 (35.51\%$\downarrow$)&24.98 (39.6\%$\downarrow$)&91.28\%\\
        HT-ResNet-20 (3-path)&199,898 (26.64\%$\downarrow$)&27.42 (33.6\%$\downarrow$)&91.29\%\\
        HT-ResNet-20 (4-path)&224,090 (17.76\%$\downarrow$)&29.87 (27.7\%$\downarrow$)&91.50\%\\
        HT-ResNet-20 (5-path)&248,282 (8.88\%$\downarrow$)&32.31 (21.8\%$\downarrow$)&91.58\%\\
        HT-ResNet-20 (6-path)&272,474 (0.00\%$\downarrow$)&34.76 (15.9\%$\downarrow$)&91.21\%\\
        \bottomrule
\end{tabular}
\end{sc}
\end{small}
\end{center}
    
\begin{center}
\begin{small}
\begin{sc}
    \caption{ImageNet-1K center-crop accuracy with different input sizes.}
    \label{tab: ImageNet-1K}
    \begin{tabular}{lcccc}
    \toprule
            Method&Parameters (M)&MACs (G)&Top-1&Top-5\\
            \midrule
        ResNet-50 (TorchVision)~\cite{he2016deep} &25.56&4.12&76.13\%&92.86\%\\
        ResNet-50 (AugSkip) ZerO Init~\cite{zhao2021zero}&25.56&4.12&76.37\%&-\\
        ResNet-50 (our trial, baseline)&25.56&4.12&76.06\%&92.85\%\\
        HT-ResNet-50 (3-path)&22.63 (11.5\%$\downarrow$)&3.60 (12.6\%$\downarrow$)&76.36\%&93.02\%\\\midrule
        ResNet-50 (our trial, baseline, 256$\times$256 input)&25.56&5.38&76.18\%&92.94\%\\
        HT-ResNet-50 (3-path, 256$\times$256 input)&22.63 (11.5\%$\downarrow$)&4.58 (14.9\%$\downarrow$)&76.77\%&93.26\%\\
        \bottomrule
        \multicolumn{5}{p{0.94\linewidth}}{Note:~\cite{deveci2018energy,akhauri2019hadanets} contain no ResNet-50-based result, but all of their networks produce worse accuracy results than their baseline models according to Table 4 in each paper.}
    \end{tabular}
    \end{sc}
\end{small}
\end{center}

\begin{center}
\begin{small}
\begin{sc}
    \caption{ImageNet-1K 10-crop accuracy with different input sizes.}
    \label{tab: ImageNet-1K-10}
    \begin{tabular}{lcccc}
     \toprule
    Method&Parameters (M)&MACs (G)&Top-1&Top-5\\
    \midrule
        ResNet-50 (TorchVision)~\cite{he2016deep} &25.56&4.12&77.43\%&93.75\%\\
        ResNet-50 (our trial, baseline)&25.56&4.12&77.53\%&93.75\%\\
        HT-ResNet-50 (3-path)&22.63 (11.5\%$\downarrow$)&3.60 (12.6\%$\downarrow$)&77.79\%&94.02\%\\
        \midrule
        ResNet-50 (our trial, baseline, 256$\times$ 256 input)&25.56&5.38&77.61\%&93.88\%\\
        HT-ResNet-50 (3-path, 256$\times$ 256 input)&22.63 (11.5\%$\downarrow$)&4.58 (14.9\%$\downarrow$)&78.33\%&94.14\%\\
        \bottomrule
    \end{tabular}
    \end{sc}
\end{small}
\end{center}
    \end{table*}
    
\begin{figure*}[htbp]
\vskip 0.2in
\begin{center}
\subfloat[The input size is $224\times 224$.]{\includegraphics[width=0.4\linewidth]{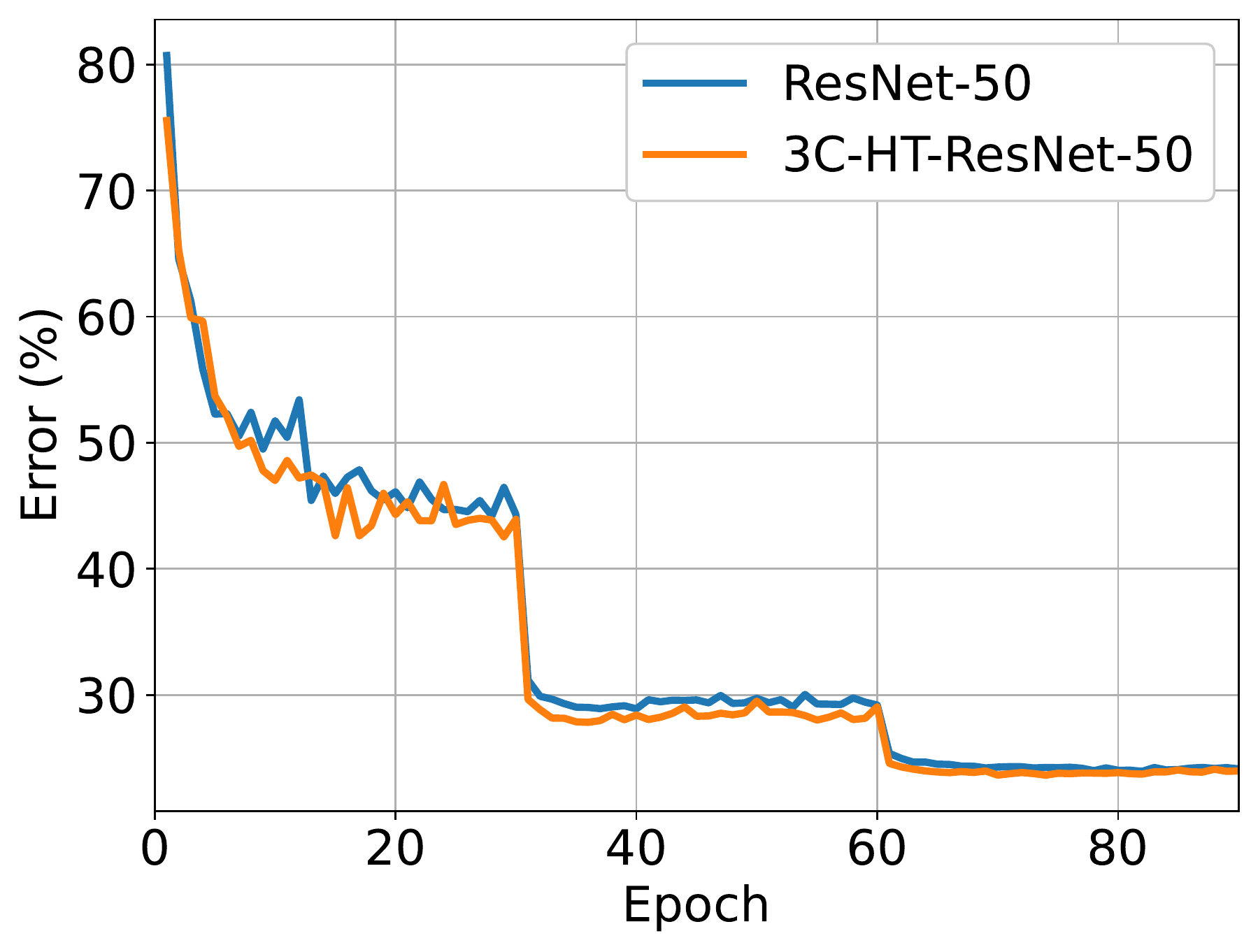}}
\subfloat[The input size is $256\times 256$.]{\includegraphics[width=0.4\linewidth]{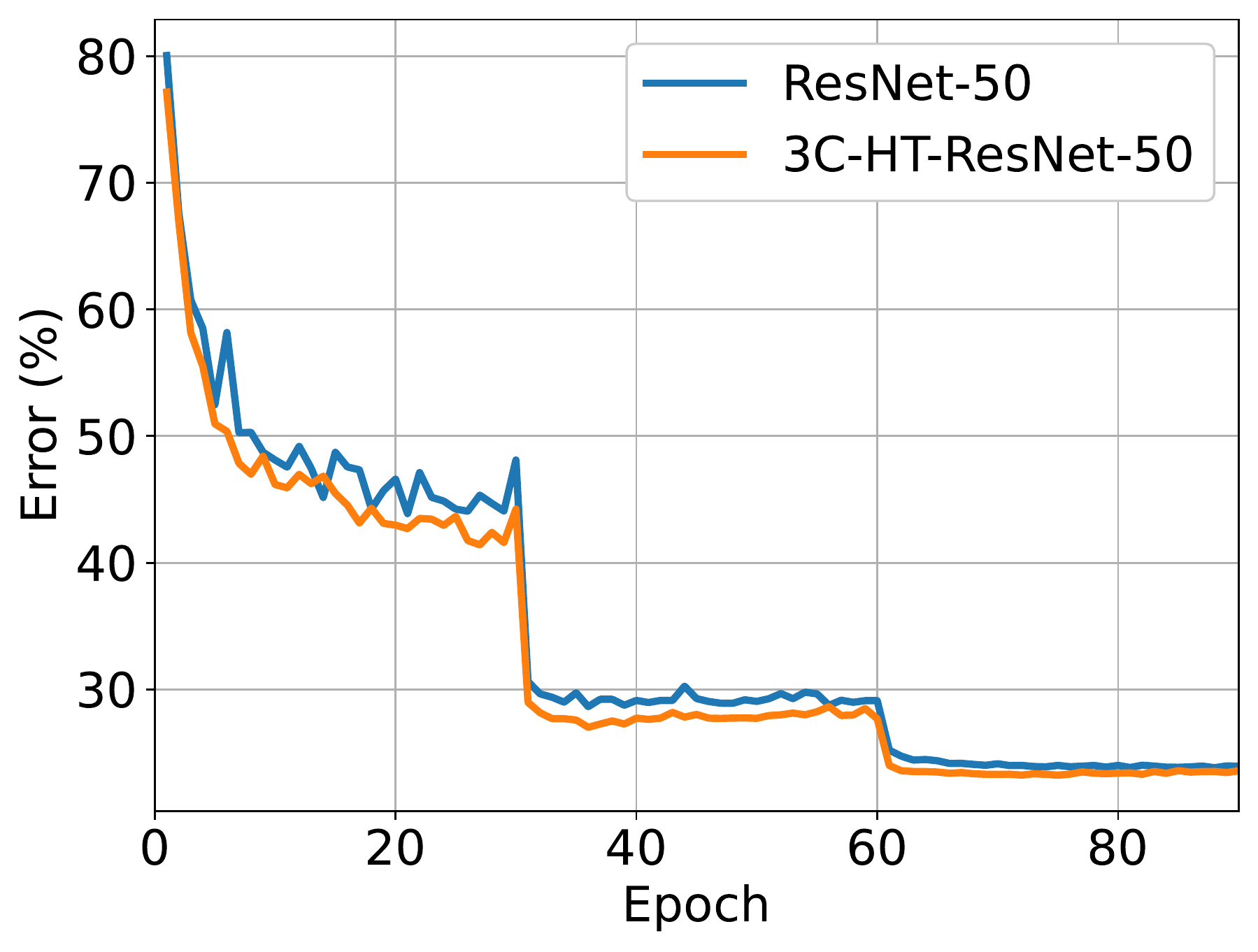}}
\caption{Training on ImageNet-1K. Curves denote the validation center-crop top-1 errors.}
\label{fig: ImageNet-1K}
\end{center}
\vskip -0.2in
\end{figure*}

In the ImageNet-1K experiments, we compare the baseline ResNet-50 model with the 3-path HT-ResNet-50. However, the most commonly used input size in the ImageNet-1K tasks is $224\times 224$, which makes the input tensors sizes not integers power of $2$. We carry out two data pre-processing approaches: First, we still use the input size of $224\times 224$, but before each HT-perceptron layer, we apply zero-padding to make the input tensor size be the integers power of $2$. This approach provides a comparison with other state-of-the-art HT-based works. Second, we use the input size of $256\times256$ instead. The number of parameters in each approach is the same, but more computational consumption is required in the second approach. As compensation, higher accuracy results are obtained using the second approach. 
We use the SGD optimizer with a weight decay of 0.0001 and momentum of 0.9. Models are trained with a mini-batch size of 128, the initial learning rate is 0.05 for 90 epochs. The learning rate is reduced by 1/10 after every 30 epochs. For data argumentation, we apply random resized crops on training images to get 224 by 224 or 256 by 256 images, then we randomly flip images horizontally. We normalize the images with the means of [0.485, 0.456, 0.406] and the standard variations of [0.229, 0.224, 0.225], respectively. We evaluate our models on the ImageNet-1K validation dataset. During the training, the best models are saved based on the center-crop top-1 accuracy on the ImageNet-1K validation dataset, and their accuracy numbers are reported in Tables~\ref{tab: ImageNet-1K} and \ref{tab: ImageNet-1K-10}. Figure~\ref{fig: ImageNet-1K} shows the center-crop top-1 error history on the ImageNet-1K validation dataset during the training phase. In the last two rows in Tables~\ref{tab: ImageNet-1K} and \ref{tab: ImageNet-1K-10} the input size is $256\times256$. In other rows, the input size is $224\times 224$. With our revision using the 3-path HT perceptron layer, $11.5\%$ parameters and more than $12\%$ MACs are reduced. In both $224\times 224$ and $256\times 256$ resolutions, the HT-ResNet-50 obtains higher accuracy results than the baseline regular ResNet-50. We think the accuracy of the HT-ResNet-50 is even better than the vanilla ResNet-50 on ImageNet is because of two reasons: First, we use the regular $3\times3$ Conv2D layer and the proposed HT-perceptron layer one after another, the HT-ResNet can extract features in different manners and fuses them. Second, we perform denoising in the transform domain similar to the classical denoising based on wavelet and other transform domain methods~\cite{bruce1994denoising}. 


\section{Conclusion}
We present a novel Hadamard Transform (HT)-based neural network layer called the HT-perceptron layer. It is a hybrid quantum-classical approach to implementing the Conv2D layer in regular CNNs. The idea is based on the HT convolution theorem which states that the dyadic convolution between two tensors is equivalent to the element-wise multiplication of their HT representations. As a result, we perform the convolutions in the HT domain using element-wise multiplications. Computing the HT is simply the application of a Hadamard gate to each qubit individually in a quantum computer. Therefore, the HT operations of the proposed layer can be implemented on a quantum computer such as an IBM-Q. Compared to the regular Conv2D layer, the proposed HT-perceptron layer is more computationally efficient as our layer requires fewer MACs than the regular Conv2D layer. We compared our proposed HT-layer-based ResNet with the regular Conv2D-based ResNets in the MNIST, CIFAR-10, and ImageNet-1K tasks, and the ResNets using the HT-perceptron layer obtain higher accuracy results on the ImageNet-1K using significantly fewer parameters and MACs than the regular networks. Our code is released at \url{https://github.com/phy710/ICML2023-HT}.

\section*{Acknowledgements}
This work was supported by National Science Foundation (NSF) under grant 1934915 and the University of Illinois Chicago Discovery Partners Institute Seed Funding Program. We thank Dr. Nhan Tran of Fermi Lab for introducing the use of Hadamard Transform in Quantum Computing to us. We sincerely appreciate all valuable comments and suggestions from ICML reviewers, which helped us in improving the quality of the paper.
\bibliography{example_paper}
\bibliographystyle{icml2023}

\newpage
\appendix
\onecolumn
\section{Proof of Lemma~\ref{lemma: HT positive}}\label{proof: HT positive}

\begin{proof}
For $k=0, 1, \ldots, N-1$, $X_k$ is the form of a constant multiple of $x_0\pm x_1\pm\ldots\pm x_{N-1}$. Therefore, if $x_0>\sum_{k=1}^{N-1}|x_k|$, then $X_k>0$ for $k=0, 1, \ldots, N-1$.
\end{proof}

\section{Proof of Theorem~\ref{Hadamard convolution theorem} Hadamard convolution theorem}\label{proof: Hadamard convolution theorem}
\begin{proof}
The sufficient condition can be proved using mathematical induction:
\begin{enumerate}
    \item It is obvious that the theorem holds for $M=0$, as with a single entry, $\mathbf{Y}=\mathbf{y}$, $\mathbf{A}=\mathbf{a}$, $\mathbf{X}=\mathbf{x}$.
    \item Suppose the theorem holds for $M\geq 0$, we prove that it also holds for $M+1$.
    Let $\mathbf{a} = [\mathbf{a}_0,\ \mathbf{a}_1]^T$ and $\mathbf{x} = [\mathbf{x}_0,\ \mathbf{x}_1]^T$, $\mathbf{a}_0, \mathbf{a}_1, \mathbf{x}_0, \mathbf{x}_1\in \mathbb{R}^{N}, N=2^M$. Then, because of Eq.~(\ref{eq: Hadamard matrix}), we have
    \begin{equation}
        \mathbf{A} = \mathcal{H}(\mathbf{a})=\sqrt{\frac{1}{2N}}\begin{bmatrix}
					\mathbf{H}_{N} & \mathbf{H}_{N} \\ \mathbf{H}_{N} & -\mathbf{H}_{N}
				\end{bmatrix}\begin{bmatrix}
					\mathbf{a}_0\\\mathbf{a}_1
				\end{bmatrix},
    \end{equation}
    \begin{equation}
        \mathbf{X} = \mathcal{H}
    (\mathbf{x})=\sqrt{\frac{1}{2N}}\begin{bmatrix}
					\mathbf{H}_{N} & \mathbf{H}_{N} \\ \mathbf{H}_{N} & -\mathbf{H}_{N}
				\end{bmatrix}\begin{bmatrix}
					\mathbf{x}_0\\\mathbf{x}_1
				\end{bmatrix}.
    \end{equation}
   Let $\mathbf{A}_i=\mathcal{H}(\mathbf{a}_i)$ and $\mathbf{X}_i=\mathcal{H}(\mathbf{x}_i)$ for $i=0, 1$. Then,
   \begin{equation}\label{eq: A=H(a)}
        \mathbf{A} = \sqrt{\frac{1}{2}}\begin{bmatrix}
					\mathbf{A}_0+\mathbf{A}_1\\\mathbf{A}_0-\mathbf{A}_1
				\end{bmatrix},
    \end{equation}
       \begin{equation}\label{eq: X=H(x)}
        \mathbf{X} = \sqrt{\frac{1}{2}}\begin{bmatrix}
					\mathbf{X}_0+\mathbf{X}_1\\\mathbf{X}_0-\mathbf{X}_1
				\end{bmatrix}.
    \end{equation}
Using Eq.~(\ref{eq: A=H(a)}), Eq.~(\ref{eq: X=H(x)}), and the assumption that the sufficient condition holds for $M$, we can prove that if
    \begin{equation}
\begin{split}
        \mathbf{y} &= \mathbf{a}*_d\mathbf{x} = \begin{bmatrix}
					\mathbf{a}_0\\\mathbf{a}_1
				\end{bmatrix}*_d\begin{bmatrix}
					\mathbf{x}_0\\\mathbf{x}_1
				\end{bmatrix}=\begin{bmatrix}	\mathbf{a}_0*_d\mathbf{x}_0+\mathbf{a}_1*_d\mathbf{x}_1\\\mathbf{a}_0*_d\mathbf{x}_1+\mathbf{a}_1*_d\mathbf{x}_0
				\end{bmatrix},
    \end{split}
    \end{equation}
    then
    \begin{equation}
\begin{split}
\mathbf{Y}&=\mathcal{H}(\mathbf{y})=\sqrt{\frac{1}{2N}}\begin{bmatrix}
					\mathbf{H}_{N} & \mathbf{H}_{N} \\ \mathbf{H}_{N} & -\mathbf{H}_{N}
				\end{bmatrix}\begin{bmatrix}	\mathbf{a}_0*_d\mathbf{x}_0+\mathbf{a}_1*_d\mathbf{x}_1\\\mathbf{a}_0*_d\mathbf{x}_1+\mathbf{a}_1*_d\mathbf{x}_0
				\end{bmatrix}=\sqrt{\frac{1}{2}}\begin{bmatrix}					\mathbf{A}_0\circ\mathbf{X}_0+\mathbf{A}_1\circ\mathbf{X}_1+\mathbf{A}_0\circ\mathbf{X}_1+\mathbf{A}_1\circ\mathbf{X}_0\\
\mathbf{A}_0\circ\mathbf{X}_0+\mathbf{A}_1\circ\mathbf{X}_1-\mathbf{A}_0\circ\mathbf{X}_1-\mathbf{A}_1\circ\mathbf{X}_0
				\end{bmatrix}\\
&=\sqrt{\frac{1}{2}}\begin{bmatrix}
(\mathbf{A}_0+\mathbf{A}_1)\circ(\mathbf{X}_0+\mathbf{X}_1)\\
(\mathbf{A}_0-\mathbf{A}_1)\circ(\mathbf{X}_0-\mathbf{X}_1)
				\end{bmatrix}=\mathbf{A}\circ\mathbf{X}.
\end{split}
\end{equation}
\end{enumerate}

The necessary condition can be proved by writing part (2) of the proof for the sufficient condition backward. 
\end{proof}

\section{ResNet Structures Used in Experiments}\label{Figures and Tables}

\begin{table}[H]
\begin{center}
\begin{small}
\begin{sc}
\caption{Revising ResNet-20 for the CIFAR-10 classification task. HT-P stands for the proposed HT-perceptron layer. Building blocks are shown in brackets, with the numbers of blocks stacked. Downsampling is performed by Conv3\_1 and Conv4\_1 with a stride of 2.}
\label{tab: resnet-20}
\vskip 0.15in
\centering
    \begin{tabular}{lcc}
    \toprule
		Layer&Output Shape&Implementation\\
        \midrule
		Input&$3\times32\times32$&-\\
		Conv1&$16\times32\times32$&$3\times3, 16$\\
		Conv2\_x&$16\times32\times32$&$\left[ \begin{array}{c} 3\times3, 16  \\ \text{HT-P}, 16 \end{array}\right]\times 3$\\
		Conv3\_x&$32\times16\times16$&$\left[ \begin{array}{c} 3\times3, 32  \\ \text{HT-P}, 32 \end{array}\right]\times 3$\\
		Conv4\_x&$64\times8\times8$&$\left[ \begin{array}{c} 3\times3, 32  \\ \text{HT-P}, 64 \end{array}\right]\times 3$\\
		GAP&$64$&Average Pooling\\
		Output&$10$&Linear, $10$\\
        \bottomrule
	\end{tabular}
 \end{sc}
\end{small}
\end{center}
\vskip -0.1in
\end{table}

\begin{table}[tb]
\begin{center}
\begin{small}
\begin{sc}
\caption{Revising ResNet-50 for the ImageNet-1K classification task. HT-P stands for the proposed HT-perceptron layer.}
\label{tab: resnet-50}
\vskip 0.15in
    \begin{tabular}{lcc}
    \toprule
		\bf{Layer}&\bf{Output Shape}&\bf{Implementation Details}\\
        \midrule
		Input&$3\times224\times224$&-\\
		Conv1&$64\times112\times112$&$7\times7, 64$, stride 2\\
		MaxPool&$64\times56\times56$&$2\times2$, stride 2\\
		Conv2\_1&$256\times56\times56$&$\left[ \begin{array}{c} 1\times1, 64  \\ 3\times3, 64, \\ 1\times1, 256 \end{array}\right]$\\
		Conv2\_2&$256\times56\times56$&$\left[ \begin{array}{c} 1\times1, 64  \\ \text{HT-P}, 64\\ 1\times1, 256 \end{array}\right]$\\
      Conv2\_3&$256\times56\times56$&$\left[ \begin{array}{c} 1\times1, 64  \\ 3\times3, 64 \\ 1\times1, 256 \end{array}\right]$\\
		Conv3\_1&$512\times28\times28$&$\left[ \begin{array}{c} 1\times1, 128  \\ 3\times3, 128, \text{stride }2 \\ 1\times1, 512 \end{array}\right]$\\
		Conv3\_2&$512\times28\times28$&$\left[ \begin{array}{c} 1\times1, 128  \\ \text{HT-P}, 128 \\ 1\times1, 512 \end{array}\right]$\\
  Conv3\_3&$512\times28\times28$&$\left[ \begin{array}{c} 1\times1, 128  \\ 3\times 3, 128 \\ 1\times1, 512 \end{array}\right]$\\
  Conv3\_4&$512\times28\times28$&$\left[ \begin{array}{c} 1\times1, 128  \\ \text{HT-P}, 128 \\ 1\times1, 512 \end{array}\right]$\\
		Conv4\_1&$1024\times14\times14$&$\left[ \begin{array}{c} 1\times1, 256  \\ 3\times3, 256, \text{stride }2 \\ 1\times1, 1024 \end{array}\right]$\\
		Conv4\_2&$1024\times14\times14$&$\left[ \begin{array}{c} 1\times1, 256  \\ \text{HT-P}, 256 \\ 1\times1, 1024 \end{array}\right]$\\
  Conv4\_3&$1024\times14\times14$&$\left[ \begin{array}{c} 1\times1, 256  \\ 3\times 3, 256 \\ 1\times1, 1024 \end{array}\right]$\\
  Conv4\_4&$1024\times14\times14$&$\left[ \begin{array}{c} 1\times1, 256  \\ \text{HT-P}, 256 \\ 1\times1, 1024 \end{array}\right]$\\
  Conv4\_5&$1024\times14\times14$&$\left[ \begin{array}{c} 1\times1, 256  \\ 3\times 3, 256 \\ 1\times1, 1024 \end{array}\right]$\\
  Conv4\_6&$1024\times14\times14$&$\left[ \begin{array}{c} 1\times1, 256  \\ \text{HT-P}, 256 \\ 1\times1, 1024 \end{array}\right]$\\
		Conv5\_1&$2048\times7\times7$&$\left[ \begin{array}{c} 1\times1, 512  \\ 3\times3, 512, \text{stride }2 \\ 1\times1, 2048 \end{array}\right]$\\		
		Conv5\_2&$2048\times7\times7$&$\left[ \begin{array}{c} 1\times1, 512  \\ \text{HT-P}, 512 \\ 1\times1, 2048 \end{array}\right]$\\	
  Conv5\_3&$2048\times7\times7$&$\left[ \begin{array}{c} 1\times1, 512  \\ 3\times3, 512 \\ 1\times1, 2048 \end{array}\right]$\\	
		GAP&$2048$&Global Average Pooling\\
		Output&$1000$&Linear, $1000$\\
            \bottomrule
	\end{tabular}
 \end{sc}
\end{small}
\end{center}
\vskip -0.1in
\end{table}


\end{document}